\pdfoutput=1
\documentclass{article}

\usepackage{microtype}
\usepackage{graphicx}
\usepackage{subfigure}
\usepackage{booktabs} % for professional tables

\usepackage{hyperref}

\usepackage[accepted]{icml2023}

\usepackage{amsmath}
\usepackage{amssymb}
\usepackage{mathtools}
\usepackage{amsthm}

\usepackage[capitalize,noabbrev]{cleveref}

\usepackage{wrapfig}
\usepackage{xspace}
\usepackage{multirow}
\usepackage{diagbox}
\usepackage{pifont}

\newcommand{\name}[0]{FusionRetro\xspace}

\usepackage[capitalize,noabbrev]{cleveref}

\theoremstyle{plain}

\theoremstyle{definition}

\theoremstyle{remark}

\usepackage[textsize=tiny]{todonotes}

\icmltitlerunning{FusionRetro: Molecule Representation Fusion via In-Context Learning for Retrosynthetic Planning}

\begin{document}

\twocolumn[
\icmltitle{FusionRetro: Molecule Representation Fusion via In-Context Learning for Retrosynthetic Planning}



\icmlsetsymbol{equal}{*}

\begin{icmlauthorlist}
\icmlauthor{Songtao Liu}{psu}
\icmlauthor{Zhengkai Tu}{equal,mit}
\icmlauthor{Minkai Xu}{equal,stanford}
\icmlauthor{Zuobai Zhang}{equal,mila,mon}\\
\icmlauthor{Lu Lin}{psu}
\icmlauthor{Rex Ying}{yale}
\icmlauthor{Jian Tang}{mila,hec,cifar}
\icmlauthor{Peilin Zhao}{tencent}
\icmlauthor{Dinghao Wu}{psu}
\end{icmlauthorlist}

\icmlaffiliation{psu}{Pennsylvania State University}
\icmlaffiliation{stanford}{Stanford University}
\icmlaffiliation{mit}{Massachusetts Institute of Technology}
\icmlaffiliation{yale}{Yale University}
\icmlaffiliation{mila}{Mila - Québec AI Institute}
\icmlaffiliation{mon}{Université de Montréal}
\icmlaffiliation{hec}{HEC Montréal}
\icmlaffiliation{cifar}{CIFAR AI Chair}
\icmlaffiliation{tencent}{Tencent AI Lab}

\icmlcorrespondingauthor{Songtao Liu}{skl5761@psu.edu}

\icmlkeywords{Machine Learning, ICML}

\vskip 0.3in
]



\printAffiliationsAndNotice{\icmlEqualContribution} % otherwise use the standard text.

\begin{abstract}
Retrosynthetic planning aims to devise a complete multi-step synthetic route from starting materials to a target molecule. Current strategies use a decoupled approach of single-step retrosynthesis models and search algorithms, taking only the product as the input to predict the reactants for each planning step and ignoring valuable context information along the synthetic route. In this work, we propose a novel framework that utilizes context information for improved retrosynthetic planning. We view synthetic routes as reaction graphs and propose to incorporate context through three principled steps: \textit{encode} molecules into embeddings, \textit{aggregate} information over routes, and \textit{readout} to predict reactants. Our approach is the first attempt to utilize in-context learning for retrosynthesis prediction in retrosynthetic planning. The entire framework can be efficiently optimized in an end-to-end fashion and produce more practical and accurate predictions. Comprehensive experiments demonstrate that by fusing in the context information over routes, our model significantly improves the performance of retrosynthetic planning over baselines that are not context-aware, especially for long synthetic routes. Code is available at \url{https://github.com/SongtaoLiu0823/FusionRetro}.

\end{abstract}

\section{Introduction}
\begin{figure}[t]
\centering
\includegraphics[width=0.37\textwidth]{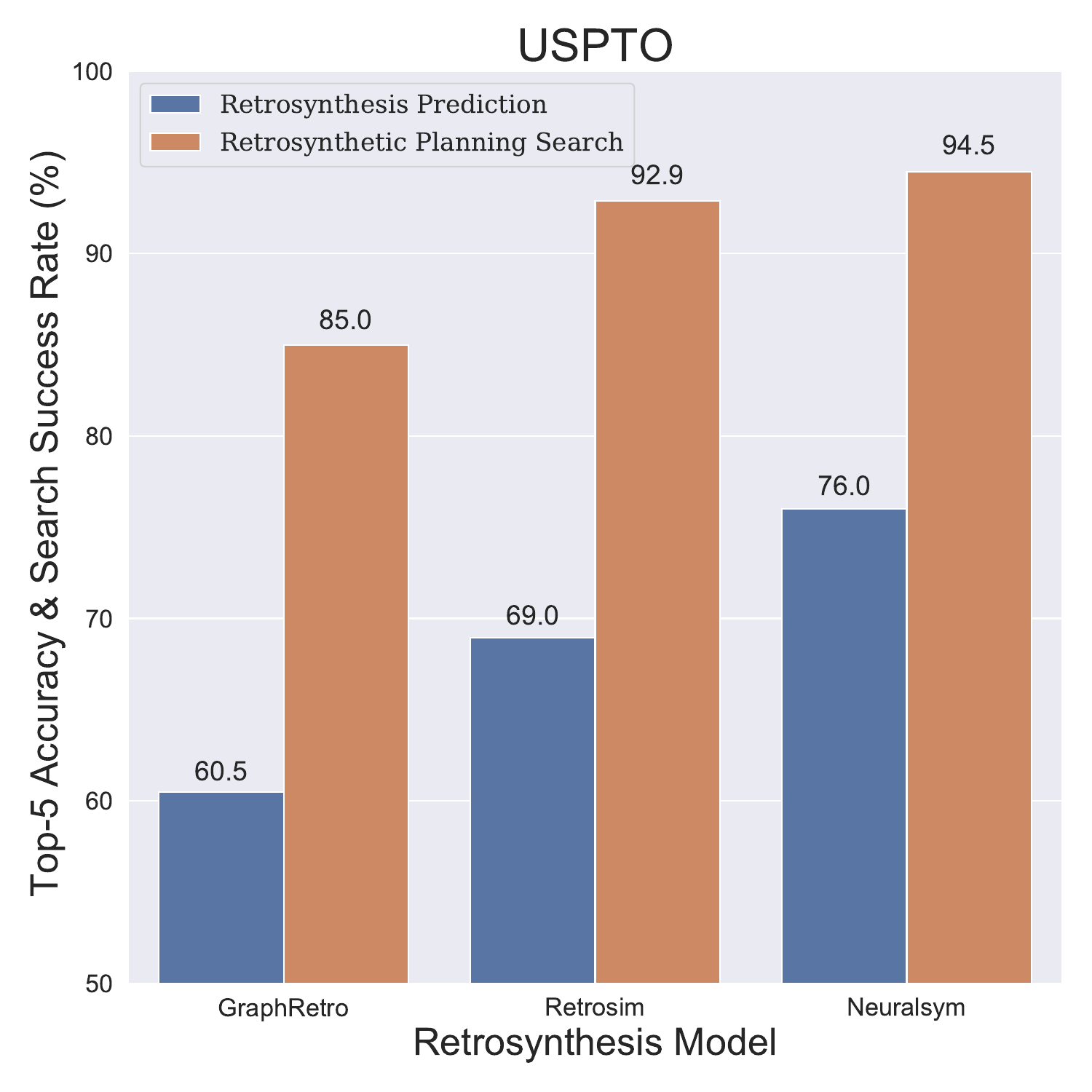}
\vskip -0.1in
\caption{Performance of retrosynthesis prediction and multi-step planning on USPTO dataset. We report the search success rate of retrosynthesis models combined with Retro* at the limit of 500 calls and 5 expansions. The search success rate is much higher than the accuracy of the top-5 retrosynthesis prediction.}
\vskip -0.1in
\label{fig:retro_search}
\end{figure}
Retrosynthetic planning is a fundamental problem in organic chemistry~\citep{coley2018machine,genheden2020aizynthfinder}. The goal of retrosynthetic planning is to find a viable set of starting materials and a sequence of reactions, that lead to a given target molecule. It is crucial for process chemistry, which aims to design efficient routes to synthesize desired target products at a low cost, as well as for materials and molecule discoveries that are contingent on the targets being synthesizable. In the past few years, with the advancement in deep learning, there has been increasing interest in applying machine learning to retrosynthetic planning, a sub-topic of Computer-Aided Synthesis Planning (CASP).

Existing strategies~\citep{segler2018planning,kishimoto2019depth,chen2020retro,lin_automatic_2020,Schwaller_2020_Hypergraph,kim2021self,xie2022retrograph,yu2022grasp} generally model retrosynthetic planning as a search problem. In a typical formulation, the synthetic route is treated as a tree or a graph, and the molecules as nodes. Starting from the target as the root node, these approaches employ some (possibly learned) search algorithms to select the most promising node to expand, and then expand it into reaction precursors with a one-step retrosynthesis model, until a viable route is found in which all the leaf nodes are commercially available.

It was not until recently that the evaluation criteria for multi-step search have somewhat converged to a few. One of the most heavily used metrics is the success rate of finding a viable route given an iteration limit (generally up to $500$). However, the search success rate is overly lenient without checking whether the searched set of starting materials can go through a sequence of reactions to synthesize the target molecule at all. This is especially problematic for the targets requiring long routes to synthesize, in which case the errors can multiply. As a quantitative illustration in Figure~\ref{fig:retro_search}, when we combine existing one-step models with top-5 accuracies between $60$ and $80$ percents with Retro*~\citep{chen2020retro}, an established search algorithm, the search success rates easily reach over $85$ and $94$ percents respectively. This is counterintuitive as we would expect the route to be less likely to succeed as more synthesis steps are added, which also throws concerns about the quality of proposed routes that the multi-step planner deems as ``successful''.

In this work, we therefore introduce the set-wise exact match of proposed starting materials to the ground truth as an alternative metric that better reflects reality. The underlying assumption is that if we can get the set of building blocks right, recovering all the reactions that ultimately lead to the target is a much easier process, possibly with the help of powerful reaction outcome predictors that can have a high accuracy of more than $90$\%~\citep{irwin2022chemformer,tetko2020state}. We construct a new benchmark with 58,099 synthetic routes retrieved from the public USPTO dataset for evaluation, in which we study the performance of multiple single-step retrosynthesis models in the context of multi-step planning, an important comparison that has not yet been done to the best of our knowledge. In addition, with our new framework of evaluation, it is now possible to consider and thereby improve the performance of all pieces in the CASP workflow in an integrated and holistic manner.

Under this view, it becomes immediately apparent that a missed opportunity by previous work is the explicit modeling of the contextual information of in-context reactions along the partial synthetic routes preceding any given node, which we subsequently explore. We propose a novel and principled context-aware model by fusing in the context embeddings, named \textbf{\name}, which is the first attempt to exploit in-context learning~\citep{min-etal-2022-metaicl} for retrosynthesis prediction in retrosynthetic planning. Specifically, we view the synthetic routes as reaction graphs and formulate our model as an end-to-end framework which: 1) \textbf{encodes} molecules on the synthetic routes into embeddings through molecule encoders; 2) \textbf{aggregates} the embeddings of molecules on the synthetic route (reaction graph) by message passing and fuses in the representations of informative contexts; and 3) \textbf{readouts} to predict the reactants on the current retrosynthetic step based on both the product and context representations learned in the previous stage.

Extensive experimental results on retrosynthetic planning tasks show that \name can achieve significantly better performance over template-free baselines, with up to a 6\% improvement in top-1 test accuracy. The surprisingly superior performance demonstrates the effectiveness of exploiting the context information and opens up room for future research in this direction.

Our contribution can be summarized as follows:
\begin{itemize}
\item We introduce a new evaluation protocol for assessing the performance of single-step retrosynthesis models in the context of multi-step planning, and for this purpose, we curate a new benchmark dataset. Our empirical analysis confirms the pivotal role of single-step accuracy in multi-step planning.

\item We propose a novel fusion framework that enables single-step models to leverage the contextual information in the reaction graph. Our method serves as the pioneering effort to exploit in-context learning~\citep{dong2022survey} for solving scientific problems, which led to impressive success in recent large language models such as ChatGPT~\citep{brown2020language}.

\item Extensive experimental results demonstrate that our proposed module can enhance the performance of the baseline model noticeably, providing insightful guidance for future research in this direction.
\end{itemize}

\section{Related Work}
\paragraph{Single-step Retrosynthesis Model.} 
Existing machine learning approaches for single-step retrosynthesis prediction can be classified into template-based and template-free models based on whether they rely on the use of reaction templates. 

Template-based algorithms~\citep{chen2020retro,coley2017computer,dai2019retrosynthesis,segler2017neural,chen2021deep,seidl2021modern} first extract these patterns from the training data, and then formulate the task as template classification or template retrieval. One of the intrinsic limitations of template methods is the need to find the right level of specificity for template definition so that they can capture sufficient chemical information without being overly specific to any reaction. As a remediation, researchers have come up with template-free methods, which have become more and more popular recently.

Template-free approaches generally use an end-to-end translation-based~\citep{liu2017retrosynthetic,zheng2019predicting,chen2019learning,karpov2019transformer,sun2021towards} or a graph-edit based formulation~\citep{sacha2021molecule}. The former models the product-to-reactants transformation as a sequence-to-sequence task by representing molecules with SMILES string, and the latter as a sequence of graph edits to atoms and bonds. 

A special family of template-free methods, which are commonly referred to as semi-template-based methods~\citep{shi2020graph,yan2020retroxpert,somnath2021learning}, adopts a two-stage formulation to first identify the reaction center(s). The target is subsequently broken into several disconnected subgraphs (i.e., the synthons), based on which the full molecule structures of reactants are recovered either by attaching the leaving group~\citep{somnath2021learning} or by generative modeling~\citep{shi2020graph,yan2020retroxpert}. For a more comprehensive understanding of the retrosynthesis literature, readers are encouraged to refer to the survey paper~\citep{meng2023retro}.

\paragraph{Search Algorithm in Retrosynthetic Planning.} 
Existing deep learning-based CASP models treat retrosynthetic planning as a search problem, which can be classified into Monte Carlo Tree Search (MCTS)~\citep{segler2018planning,hong2021retrosynthetic}, Proof-Number Search (PNS)~\citep{kishimoto2019depth}, A*-like Search~\citep{chen2020retro,han2022gnn,xie2022retrograph}, and Reinforcement Learning (RL) based Search~\citep{yu2022grasp}. \citet{segler2018planning} integrates MCTS with policy networks to guide multi-step planning. Drawing inspiration from search techniques in two-player zero-sum games, DFPN-E~\citep{kishimoto2019depth} combines Depth-First Proof-Number (DFPN) with Heuristic Edge Initialization for chemical synthesis planning. Retro*~\citep{chen2020retro} introduces a neural-based A*-like algorithm to estimate solution costs and select the most promising one. GRASP~\citep{yu2022grasp} leverages reinforcement learning to guide the search process. Both GNN-Retro~\citep{han2022gnn} and RetroGraph~\citep{xie2022retrograph} employ graph neural networks~\citep{kipf2017semi} to aggregate information from the synthetic route, thereby enabling more accurate estimation of costs in Retro*. All works thus far, however, treat the selection policy and the expansion policy (i.e., the single-step model) as two disjoint pieces. Usage of context information of partially explored synthesis trees is non-existent in the single-step predictor, and to a minimal extent in the search phase in the form of some cost functions that are updated as planning proceeds. 

In contrast, our work \emph{explicitly} integrates reactions along the synthetic routes as in-context examples into our single-step model. We achieve this by fusing in the product embedding directly into the model inputs. Compared with context-aware A*-like search algorithms~\citep{han2022gnn,xie2022retrograph}, our proposed approach, focusing on retrosynthesis prediction, is a modular framework comprising encoding, fusion, and readout components. Our fusion leverages in-context learning to maximize the use of in-context reactions. Importantly, it is not limited to GNNs alone and can incorporate various aggregation
methodologies, such as Transformer~\citep{vaswani2017attention} and Graph Transformer~\citep{ying2021transformers}. This framework lays a solid foundation for future explorations in the design of retrosynthesis models for retrosynthetic planning, specifically targeting three key aspects: encoding, fusion, and readout modules.

\paragraph{Evaluation of Retrosynthetic Planning.}
The de facto standard for evaluating single-step retrosynthesis models has been the top-k accuracy, or whether the ground truth reactants appear in the top-k suggestions. Alternatives such as accuracy for the largest predicted fragment~\citep{tetko2020state} and round-trip accuracy based on how likely the proposed reactants can lead to the product~\citep{Schwaller_2020_Hypergraph} have been proposed and sometimes used in parallel with top-k accuracy. All of these metrics solely evaluate the models in the single-step context, but how the single-step performance translates into the likelihood of success in multi-step planning remains an open question. The evaluation of multi-step planning, on the other hand, tends to have two distinct focuses, either on \emph{efficiency} or on \emph{quality}. Search efficiency has been measured in the success rate of finding pathways with buyable starting materials, as well as average numbers of iterations and node visits. However, as we demonstrate in Figure~\ref{fig:retro_search}, efficiency metrics like the success rate give little insight into route quality. To evaluate route quality, simple proxies such as route length~\citep{chen2020retro,kishimoto2019depth} and average complexity of molecules~\citep{shibukawa2020compret} have been used, and so have more complicated heuristics such as tree edit distance to a reference route~\citep{genheden2022paroutes}. 

However, some existing benchmarks~\citep{chen2020retro,genheden2022paroutes,tripp2022re} based on these metrics do not verify if the searched materials can synthesize the target molecule. Although the ideal method for validating the feasibility of starting materials would involve chemical laboratory testing or expert evaluation, these approaches are frequently cost-prohibitive. Consequently, we introduce a complementary matching metric to evaluate retrosynthesis models and search algorithms by comparing predicted starting materials with those retrieved from the dataset during testing. The set-wise exact match of starting materials, as we propose in this work, finds a balance between simplicity and data awareness. It is cheap to compute, easy to implement, and yet provides a good indication of how probable the suggested set will successfully lead to the target. Note that our introduced evaluation metric does not restrict prediction diversity. Although \citet{tripp2022re} propose a metric to evaluate prediction diversity, it does not verify whether the searched starting materials can indeed synthesize the target molecule. As such, a trade-off exists between our metric and theirs. We believe this open problem could stimulate future research to develop new metrics and methods that effectively address both aspects.

\section{Background}
\begin{figure}[t]
\centering
\includegraphics[width=0.48\textwidth,clip,trim=0 12pt 0 0]{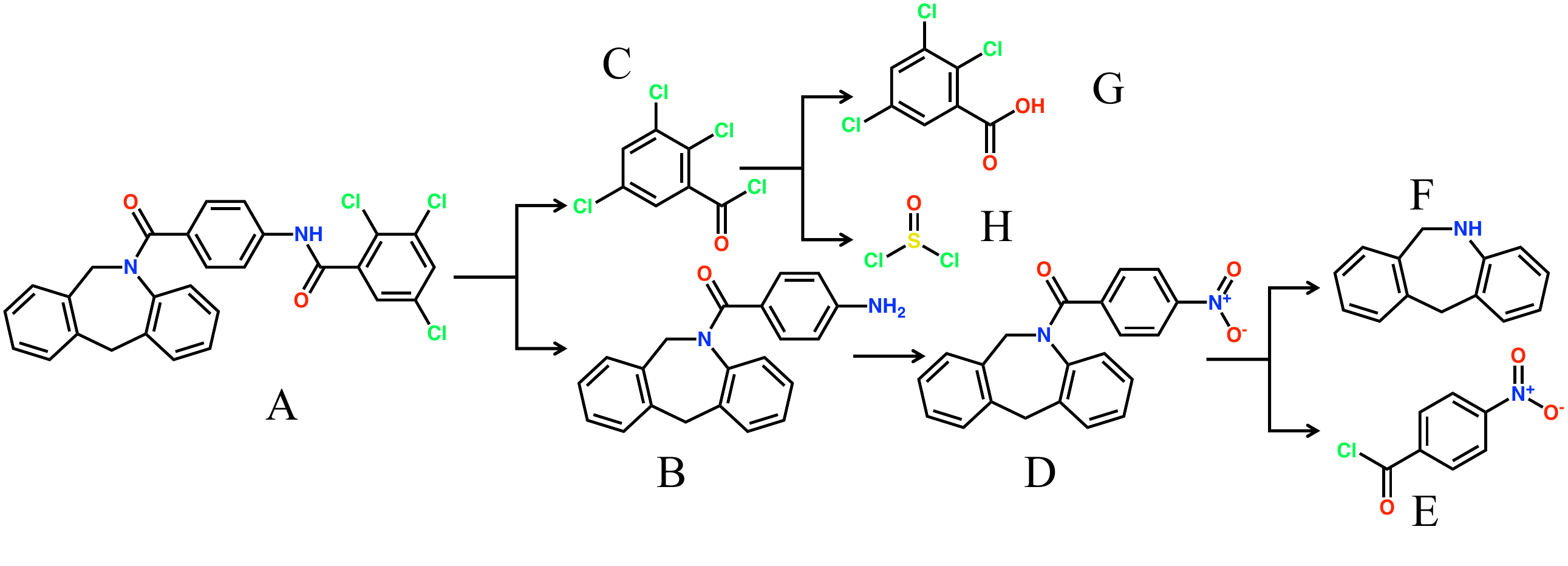}
\vskip -0.1in
\caption{Illustration of \textit{synthetic route}. Given the definition in Eq.~(\ref{def:depth}), the depth of this route is 3, which means the depth of the longest path is 3. A is the desired target molecule to be synthesized. B, C, and D are the intermediates. E, F, G, and H are the starting materials.
}
\label{fig:retro graph}
\vskip -0.1in
\end{figure}

In this section, we formally define important terminologies used in the rest of the paper, including starting materials and synthetic routes, based on which we define the formulation of retrosynthetic planning.

\paragraph{Starting Material.} 
We denote the space of all chemical molecules as $\mathcal{M}$. Following AiZynthFinder~\citep{genheden2020aizynthfinder}, we define the starting materials as a set of commercially purchasable molecules, denoted as $\mathcal{S} \subseteq \mathcal{M}$. ZINC~\citep{sterling2015zinc} releases the open-source databases of purchasable compounds. We define this list of compounds in these databases as our starting materials.

\paragraph{Synthetic Route.} 
Given the above definitions, a synthetic route can also be organized as a graph-like structure, called reaction graph~\cite{shibukawa2020compret,nguyen2021generative}. In the rest of the paper, we use the terminology ``reaction graph'' instead of ``synthetic route''. An illustration of a reaction graph (denoted as $\mathcal{G}$) is shown in Figure~\ref{fig:retro graph}. Here, $\mathcal{G}=\{T, \mathcal{R}, \mathcal{I}, \tau\}$, where $T \in \mathcal{M}\setminus \mathcal{S}$ is the target molecule we desire to synthesize (A in Figure~\ref{fig:retro graph}), $\mathcal{R}=\{r_1, r_2, \ldots, r_n\}\subseteq \mathcal{S}$ is the set of starting materials (E, F, G, H in Figure~\ref{fig:retro graph}) that go through a series of reactions $\tau$ to synthesize A, and $\mathcal{I}=\{m_1, m_2, \ldots, m_u\}\subseteq \mathcal{M}\setminus\mathcal{S}$ is the set of intermediates (B, C, D in Figure~\ref{fig:retro graph}) formed from molecules represented by their child nodes, which can react further to produce the molecule represented by their parent nodes. A reaction graph consists of multiple paths from the target molecule to any starting material in the reaction graph. According to the definition, the number of paths is equal to the number of starting materials. We denote paths as $l$, the set of paths as $\mathcal{L} = \{l_1, l_2, \ldots, l_n\}$, and we have
\begin{equation}
    \tau = \tau_{l_1} \cup \tau_{l_2} \cup \cdots \cup \tau_{l_n},
\end{equation}
where $\tau_{l_i}$ is the set of reactions accompanying path $l_i$. As illustrated in Figure~\ref{fig:retro graph}, $A\rightarrow B\rightarrow D\rightarrow E$ is one of the paths in this graph. We denote the depth $\mathcal{D_\mathcal{G}}$ of a reaction graph as the length of the longest path in this graph, where
\begin{equation}
\label{def:depth}
    \mathcal{D_\mathcal{G}} = \max_i \mathcal{D}_{l_i}.
\end{equation}
The depth of a reaction graph is also the number of steps required to synthesize a molecule from a fixed set of commercially purchasable compounds. Note that in this paper, the default order of the path is in the retrosynthetic (rather than forward) direction.

\paragraph{Single-Step Retrosynthesis.} Given a target product molecule $T \in \mathcal{M}$, the goal of one-step retrosynthesis is to predict a set of reactants $\mathcal{R}=\{r_1, r_2, \ldots, r_n\} \subseteq \mathcal{M}$ that can react to synthesize this product, which can be formulated as:
\begin{equation*}
    T \rightarrow \mathcal{R}.
\end{equation*}
\paragraph{Retrosynthetic Planning.}
Given a target molecule $T \in \mathcal{M}$, the goal of retrosynthetic planning is to search for the starting materials $\mathcal{R}=\{r_1, r_2, \ldots, r_n\} \subseteq \mathcal{S}$ that can synthesize the target molecule through a set of chemical reactions $\tau=\{R_1, R_2, \ldots, R_m\}$, which can be formulated as follows: 
\begin{equation}
    T \rightarrow \mathcal{I} \rightarrow \mathcal{R},
\end{equation}
where $\mathcal{I} \subseteq \mathcal{M}\setminus \mathcal{S}$ is the set of intermediates.

\section{FusionRetro}
\begin{figure*}[t]
\centering
\includegraphics[width=0.87\textwidth]{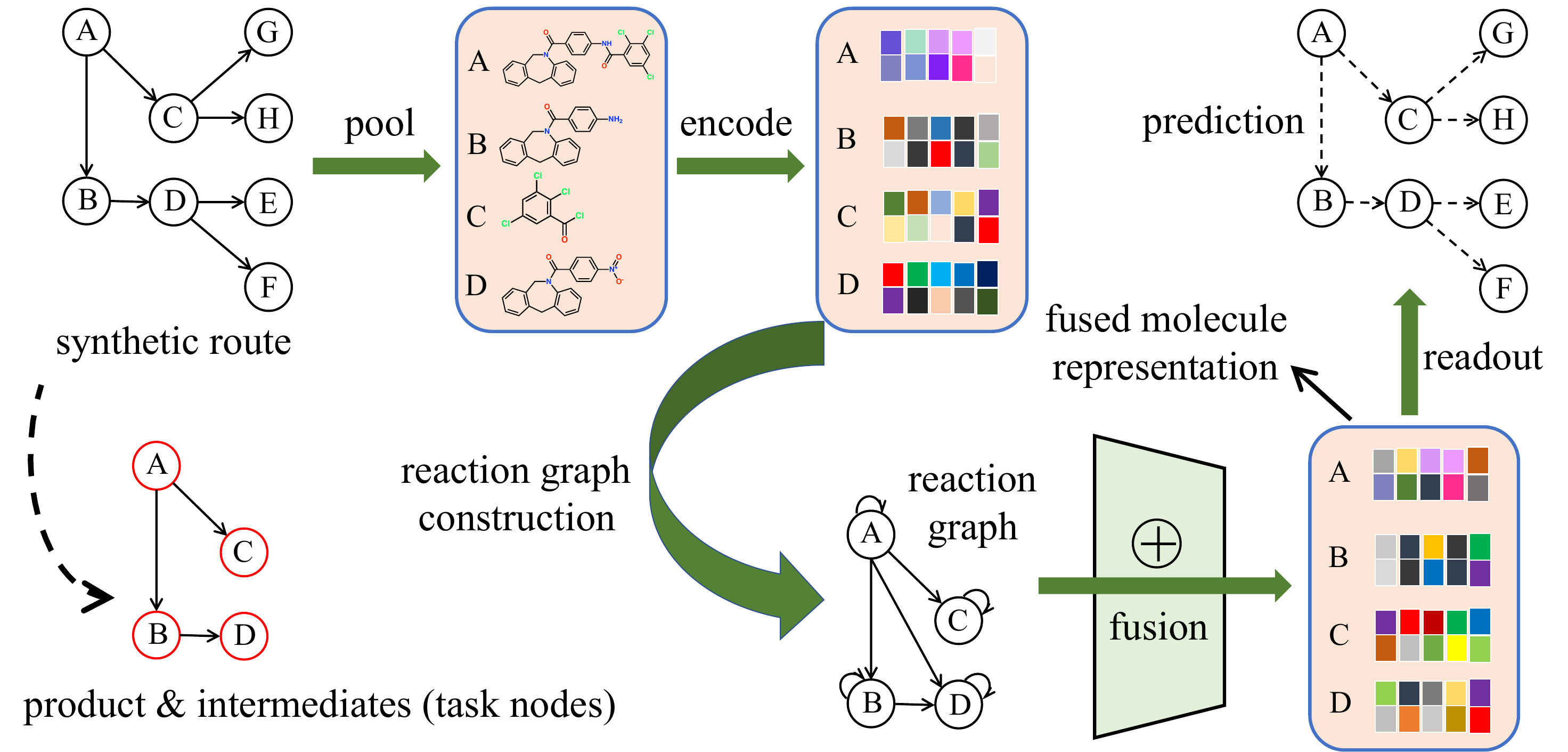}
\vskip -0.1in
\caption{Illustration of our framework. Our framework consists of three modules: \emph{encode}, \emph{aggregation}, and \emph{readout}. The process begins with the construction of a reaction graph from the given synthetic route. After encoding the molecules present in this reaction graph, we utilize the aggregation module to generate the fused molecule representations (FMR). This FMR is used for retrosynthesis prediction.}
\label{fig:framework}
\vskip -0.1in
\end{figure*}

In this section, we delve into the specifics of our proposed \name method. We commence by describing how we construct our reaction graphs in Section~\ref{ssec:method:reaction-graph}, drawing upon the synthetic routes depicted in Figure~\ref{fig:retro graph}. We then proceed to elaborate on our systematic approach for utilizing informative in-context examples (reactions) from the reaction graph in Section~\ref{ssec:method:mol-fusion}. This framework involves three principled steps: \textit{encode} molecules into embeddings, \textit{aggregate} the embeddings of molecules through message passing over reaction graphs, and \textit{readout} to predict reactants on current retrosynthesis step. We conclude this section by briefly outlining the practical aspects of our training and inference algorithm in Section~\ref{ssec:method:train-inference}. Figure~\ref{fig:framework} offers a high-level visual representation of our proposed framework.

\subsection{Reaction Graph}
\label{ssec:method:reaction-graph}
In this section, we describe the details of how to construct the reaction graph from the synthetic route.
\paragraph{Task Nodes.}
First, we introduce the concept of task molecules, which serve as the nodes in our reaction graphs. Specifically, we designate the target molecule $T$ and intermediates $\mathcal{I}$ as task molecules, as these will be expanded during the multi-step planning process. Importantly, because our search process halts when the molecules on the synthetic route are commercially available, these leaf nodes in Figure~\ref{fig:retro graph} are not included in our constructed reaction graph.

\paragraph{Graph Construction.}
In order to explicitly model the contextual information of reactions and intermediates along the synthetic routes, we build reaction graphs among task molecules. We first remove non-task molecules on the leaf nodes in Figure~\ref{fig:retro graph}. Then, inspired by the dense connection~\citep{huang2017densely} between tokens in Transformer~\citep{vaswani2017attention}, we link each task molecule and its ancestors to construct our reaction graph, which enables us to explicitly model the relational information between task molecules.

\subsection{Molecule Representation Fusion}
\label{ssec:method:mol-fusion}
As depicted in Figure~\ref{fig:framework}, a given path consists of several chemical reactions. Inspired by the recent advancements in in-context learning within large language models, we utilize in-context examples—specifically, the reactions preceding the current one—to boost the accuracy of our current prediction. To this end, we propose a well-founded fusion framework. This framework is designed to regulate the information flow and distill representations that seize essential contextual information from the reaction graph.

Another part of our motivation stems from the discrepancy between machine learning methods prevalent in existing works and the actual thought process of chemists. Chemists don’t typically think like a search engine – by iteratively applying some rigid one-step expansion with some search criteria. Instead, many of them think in a more holistic way, for example, by taking into account all the intermediate steps when planning the next, for reasons including but not limited to ease of purification. A purely one-step model would likely miss most, if not all, of this contextual information. Thus, in this section, we delve into the specifics of molecule representation fusion. Particularly, we employ the attention mechanism to generate representations that capture the contextual information of reactions and intermediates along the reaction graph.

\begin{figure*}[t]
\centering
\includegraphics[width=0.93\textwidth]{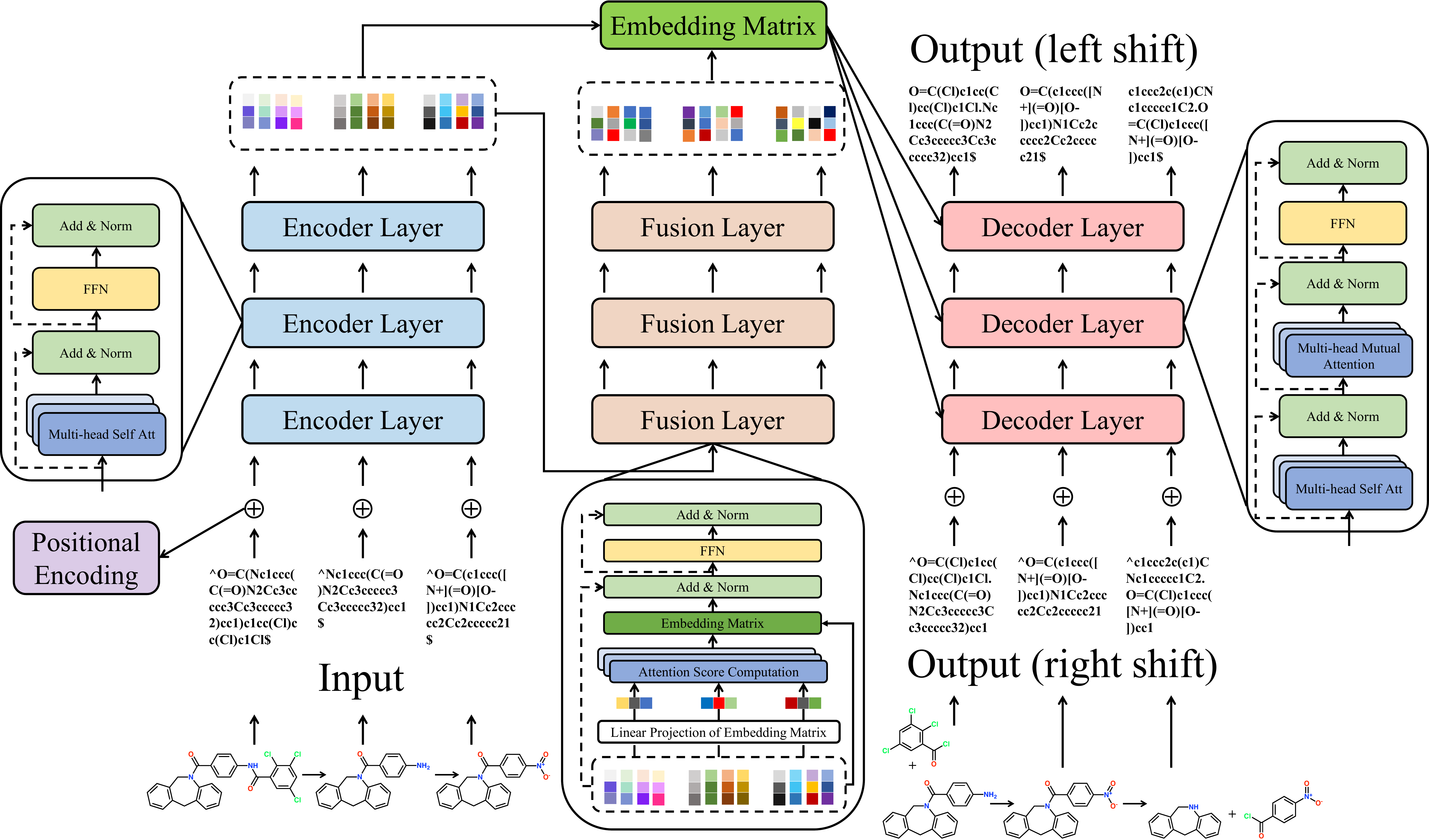}
\vskip -0.1in
\caption{Illustration of our architecture. Our architecture consists of the encoder, decoder, and fusion modules, each of which is composed of several stacked attention layers. In the encoder, we employ self-attention layers to transform the embeddings of input SMILES into latent representations, known as encoder outputs. Subsequently, we utilize the fusion module to attain a fused molecule representation. This fused representation is then fed into the decoder, which yields the final prediction.}
\vskip -0.1in
\label{arc}
\end{figure*}

\paragraph{Molecule Encoding.}
Given the reaction graph depicted in Figure~\ref{fig:framework}, the first step involves encoding the molecules in the reaction graph into embeddings using molecule encoders. These encoders can be broadly categorized into sequence-based and graph-based methods. Graph-based models~\citep{shi2020graph,yan2020retroxpert,somnath2021learning,sacha2021molecule} employ a Message Passing Neural Network~\citep{gilmer2017neural} to translate the molecule graph into an embedding vector. On the other hand, sequence-based models~\citep{karpov2019transformer} leverage the attention mechanism to transform the SMILES representation of the molecule into an embedding matrix. Note that our proposed method is a general framework, and we intentionally omit the details of the encoding process. Instead, we represent the encoder as a function, denoted as $\phi$. Therefore, the encoding process can be formulated as follows:
\begin{equation}
    \boldsymbol{h}_m = \phi(m),
\end{equation}
where $m\in\{T\}\cup \mathcal{I}$ and $\boldsymbol{h}_m$ denotes the representation of molecule $m$.
\paragraph{Representation Fusion.}
Upon encoding, we carry out a message-passing operation to aggregate the molecule embeddings. This allows us to create fused representations that encapsulate contextual information. Rather than directly employing the weights in the adjacency matrix, we compute the correlation coefficient~\citep{zhang2020gnnguard} to assess the relevance between molecule nodes $u$ and $v$. Based on these correlation coefficients, we can propagate messages across the weighted reaction graph in a more meaningful manner. To quantify the correlation between two molecule nodes, we make use of an attention mechanism~\citep{velivckovic2018graph,sukhbaatar2015end,weston2015memory}, deriving the coefficients as follows:
\begin{equation}
c(\boldsymbol{h}_u, \boldsymbol{h}_v) = \boldsymbol{h}_u \odot \boldsymbol{h}_v,
\end{equation}
where $\odot$ stands for the dot product. In a manner akin to GAT~\citep{velivckovic2018graph}, we also normalize the coefficients across all neighbors using the softmax function:
\begin{equation}
\begin{split}
\alpha(\boldsymbol{h}_u, \boldsymbol{h}_v) &=\operatorname{softmax}_v\left(c(\boldsymbol{h}_u, \boldsymbol{h}_v)\right)\\
&=\frac{\exp \left(c(\boldsymbol{h}_u, \boldsymbol{h}_v)\right)}{\sum_{k \in \mathcal{N}_u} \exp \left(c(\boldsymbol{h}_u, \boldsymbol{h}_k)\right)},
\end{split}
\end{equation}
where $\mathcal{N}_u$ represents the neighborhood of molecule $u$ in the reaction graph. With this approach, we can quantify the message transmitted along the weighted reaction graph and derive the fused representation as follows:
\begin{equation}
    \boldsymbol{h}_u^{'} = \sum_{v \in \mathcal{N}_u} \alpha(\boldsymbol{h}_u, \boldsymbol{h}_v) \boldsymbol{h}_v,
\end{equation}
where $\boldsymbol{h}_u^{'}$ denotes the fused molecule representation, which captures the contextual information and thus enables more accurate retrosynthesis predictions in multi-step planning, as will be demonstrated in the experimental section.

\paragraph{Readout.}
Upon obtaining the fused molecule representation (FMR), we employ both the FMR and the original molecule representation as input to predict the reactants using the decoder. The specifics of the readout process are not discussed here, but we represent it as a function $\psi$. Thus, the readout process can be expressed as follows:
\begin{equation}
p = \psi(\boldsymbol{h}_u, \boldsymbol{h}_u^{'}),
\end{equation}
where $p$ stands for the prediction.

\paragraph{Implementation Details.}
We implement our proposed module based on Transformer~\citep{karpov2019transformer} given its use of an end-to-end training paradigm, as depicted in Figure~\ref{arc}. It's important to note that our method is a general framework and can inspire future work to incorporate our framework into other retrosynthesis models.

\subsection{Training and Inference}
\label{ssec:method:train-inference}
\paragraph{Training.}
During the training phase, we use the entire reaction graph as input, facilitating parallel computation. Given the SMILES representations of molecules (A, B, C, D) present on the reaction graph, the output should correspond to the SMILES representations of (B+C, D, G+H, E+F). Notably, while predicting B+C, input A is regarded as informative, but the information of (B, D) are not considered. Therefore, during training, the information on child molecules is excluded when making the current prediction. Thanks to the attention mechanism, our predictions for all reactions along the reaction graph are parallelized during the training phase. This can be achieved by leveraging the adjacency matrix and masking the inputs of child molecules. The loss function can be expressed as follows
\begin{equation}
    \mathcal{L}(y, p)=-\sum_{i=1}^n\sum_{j=1}^K y_{ij} \log \left(p_{ij}\right),
\end{equation}
where $y_{ij}$ and $p_{ij}$ are the predicted and ground truth values at the $j$-th position for the $i$-th target molecule sequence. In other words, the training is parallelized on all the retrosynthesis reactions in the reaction graph.
\begin{figure}[t]
\centering
\includegraphics[width=0.48\textwidth,clip,trim=0 12pt 0 0]{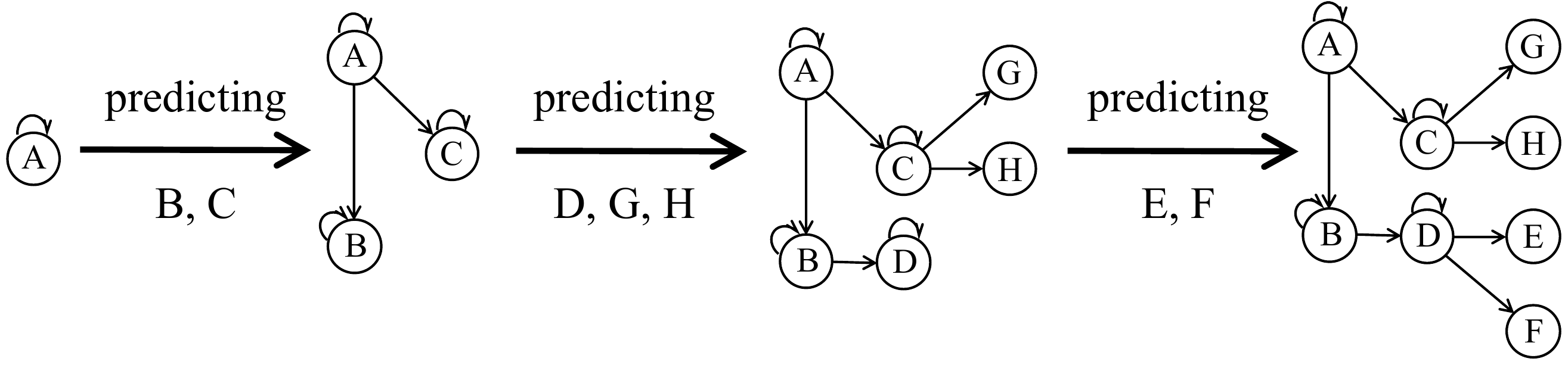}
\vskip -0.1in
\caption{Overview of the inference process. We start from the target molecule A and perform backward chaining to do a series of one-step retrosynthesis predictions until all the final reactants are starting materials. 
}
\label{fig:inference}
\end{figure}

\paragraph{Inference.}
During the inference phase, we initiate the process with the target molecule $T$ and apply backward chaining to conduct a series of one-step retrosynthesis predictions until all reactants have been identified as starting materials. After predicting the reactant molecules for each retrosynthesis step, we cross-reference the set of starting materials to verify whether the predicted reactants are indeed starting materials. If they are, we add them to the predicted reactant set. If not, we establish a new path and predict the output for the next step. The inference process concludes once the path set is emptied. This inference process is graphically depicted in Figure~\ref{fig:inference} and procedurally detailed in Algorithm~\ref{alg:inference}.

\begin{algorithm}[t]
\caption{Inference given a target molecule}
\label{alg:inference}
\begin{algorithmic}[1]
\STATE {\bfseries Input:} Target molecule $T$, starting material set $\mathcal{S}$
\STATE Initialize reactant set $\mathcal{R}=\{\}$, path set $\mathcal{L}=\{\}$
\STATE Put the initial path $[T]$ into $\mathcal{L}$
\WHILE{$\mathcal{L}$ is not a empty set}
    \STATE Take an path $l$ from $\mathcal{L}$
    \STATE Predict the reactants $r_l$ for expansion given $l$
    \FOR{reactant $r_l^{(i)}$ in $r_l$} 
        \IF{$r_l^{(i)}\in \mathcal{S}$}
            \STATE Put $r_l^{(i)}$ into $\mathcal{R}$
        \ELSE
            \STATE Generate a new path $l'=l+[r_l^{(i)}]$
            \STATE Put $l'$ into $\mathcal{L}$
        \ENDIF
    \ENDFOR
\ENDWHILE
\STATE \textbf{return} predicted reactant set $\mathcal{R}$
\end{algorithmic}
\end{algorithm}

\section{Experiments}
In this section, we evaluate the performance of different retrosynthesis models on our constructed dataset for retrosynthetic planning.

\subsection{Dataset Construction}
We construct a benchmark for retrosynthetic planning using the public USPTO-full dataset, which consists of 906,164 valid reactions from the original 1,808,937, after removing invalid and duplicate ones. These reactions are used to construct a reaction network~\citep{li2022prediction}, treating molecules with an out-degree of zero as target molecules. We use dynamic programming and backtracking to identify all synthetic routes for each target, and following the approach in \citet{chen2020retro}, we extract the shortest-possible synthetic routes with leaf nodes as starting materials. This process yields synthetic routes for 128,469 molecules. We disregard routes that synthesize target molecules in one step and split the remaining molecules into training, validation, and test datasets in an 80\%/10\%/10\% ratio. This results in 46,458 samples for training, 5,803 for validation, and 5,838 for testing. Note that the target molecules in the training set, validation set, and test set do not intersect. We call our benchmark RetroBench. Detailed statistics of the dataset can be found in Appendix~\ref{appendix:dataset}.

\begin{table*}[htbp]
\centering
\caption{Summary of retrosynthetic planning results in terms of exact match accuracy (\%).}
\label{tab:main results}
\vskip 0.1in
\setlength{\tabcolsep}{0.75mm}
\begin{tabular}{lccccccccccc}
\toprule
\multirow{2}{*}{Search Algorithm} & \multicolumn{5}{c}{Retro*} & \multicolumn{5}{c}{Retro*-0} & \multicolumn{1}{c}{Greedy DFS} \\
\cmidrule(r){2-6} \cmidrule(r){7-11} \cmidrule(r){12-12}
&  Top-1     &  Top-2  &   Top-3  &   Top-4 &   Top-5
&  Top-1     &  Top-2  &   Top-3  &   Top-4 &   Top-5
&   Top-1  \\
\midrule
\multicolumn{1}{c}{Template-based} \\
\midrule
Retrosim~\citep{coley2017computer}  &35.1  &40.5  &42.9   &44.0  &44.6  &35.0  &40.5  &43.0  &44.1 & 44.6 & 31.5 \\
Neuralsym~\citep{segler2017neural}  & \textbf{41.7}  & \textbf{49.2}  &52.1   &53.6  &54.4  &\textbf{42.0}  &\textbf{49.3}  &52.0  &53.6 & 54.3 & \textbf{39.2} \\
GLN~\citep{dai2019retrosynthesis}  &39.6  &48.9  &\textbf{52.7}   &\textbf{54.6}  &\textbf{55.7}  &39.5  &48.7  &\textbf{52.6}  &\textbf{54.5} & \textbf{55.6} & 38.0 \\
\midrule
\multicolumn{1}{c}{Template-free} \\
\midrule
G2Gs~\citep{shi2020graph}  &5.4  &8.3  &9.9   &10.9  &11.7  &4.2  &6.5  &7.6  &8.3 &8.9  & 3.8 \\
GraphRetro~\citep{somnath2021learning}  &15.3  &19.5  &21.0   &21.9  &22.4  &15.3  &19.5  &21.0 &21.9 &22.2  & 14.4 \\
Megan~\citep{sacha2021molecule}  &18.8  &29.7  &37.2   &42.6  &45.9  &19.5  &28.0  &33.2  &36.4 &38.5  &32.9  \\
Transformer~\citep{karpov2019transformer}  &31.3  &40.4  &44.7   &47.2  &48.9  &31.2  &40.5  &45.1  &47.3 &48.7  &26.7  \\
FusionRetro  & \textbf{37.5} & \textbf{45.0} & \textbf{48.2}  & \textbf{50.0} & \textbf{50.9} & \textbf{37.5}  &\textbf{45.0}  &\textbf{48.3}  &\textbf{50.2} &\textbf{51.2}  &\textbf{33.8}  \\
\bottomrule
\end{tabular}
\vskip -0.1in
\end{table*}

\subsection{Experiment Setup}
\paragraph{Evaluation Protocol.}
As previously discussed, current search algorithms~\citep{segler2018planning,kishimoto2019depth,chen2020retro,kim2021self,xie2022retrograph,yu2022grasp} primarily utilize search success rate as their evaluation metric, without verifying if the identified starting materials can indeed synthesize the target molecule. In this study, we propose a new evaluation metric: the set-wise exact match between the proposed starting materials and the ground truth. For a given target molecule, we carry out a series of one-step retrosynthesis predictions and employ search algorithms to select the most promising reactant candidates for expansion, until all leaf nodes have been identified as starting materials. We use the starting materials sourced from our constructed reaction network as the ground truth and compare them to the starting materials identified through the search. The match is based on a basic comparison of the InChiKey of the molecule, as used by AiZynthFinder~\citep{genheden2020aizynthfinder}. It's important to note that a particular target molecule may have multiple synthetic routes in the test set. We consider it an accurate match when the predicted starting material set aligns with at least one of the multiple ground truths. Additionally, we implement a pruning search, halting the search when the length of the predicted synthetic route surpasses the depth of the ground truth synthetic route. Utilizing our evaluation metric allows us to compare the performances of different retrosynthesis models in conjunction with various search algorithms, thereby providing a benchmark for future studies.

\paragraph{Setting and Baselines.}
We evaluate the effectiveness of our proposed retrosynthesis method in conjunction with three different search algorithms for retrosynthetic planning. This approach is benchmarked against existing single-step retrosynthesis models, which can be broadly categorized into two groups: template-based and template-free models. Each model is trained using the reactions in our training dataset. Upon completion of the retrosynthesis training, we employ the Retro*~\citep{chen2020retro}, Retro*-0, and Greedy DFS search algorithms. For all baselines, except for Transformer, we adhere to their original experimental setups, including hyperparameters and data processing, as described in their respective papers. These experiments are conducted using their publicly available codes. Transformer is implemented using Pytorch~\citep{paszke2019pytorch}, and we re-tuned the learning rate due to the spike phenomenon observed with the learning rate reported in the original paper. The template-based baseline approaches we consider include Retrosim~\citep{coley2017computer}, Neuralsym~\citep{segler2017neural}, and GLN~\citep{dai2019retrosynthesis}. We also evaluate end-to-end template-free approaches such as Transformer~\citep{karpov2019transformer} and Megan~\citep{sacha2021molecule}, as well as semi-template-based models like G2Gs~\citep{shi2020graph} and GraphRetro~\citep{somnath2021learning}. Our framework is depicted in Figure~\ref{arc}. For all hyperparameters, except for the learning rate (due to the spike phenomenon), we adhere to the settings reported in the publicly released Transformer code and do not perform any additional hyperparameter tuning. Detailed information on the hyperparameters can be found in Appendix~\ref{appendix:implementation}. Our proposed model, \name, is trained using 2 NVIDIA Tesla V100 GPUs.

\subsection{Results}
\begin{figure}[t]
    \begin{center}
        \includegraphics[width=0.4\textwidth]{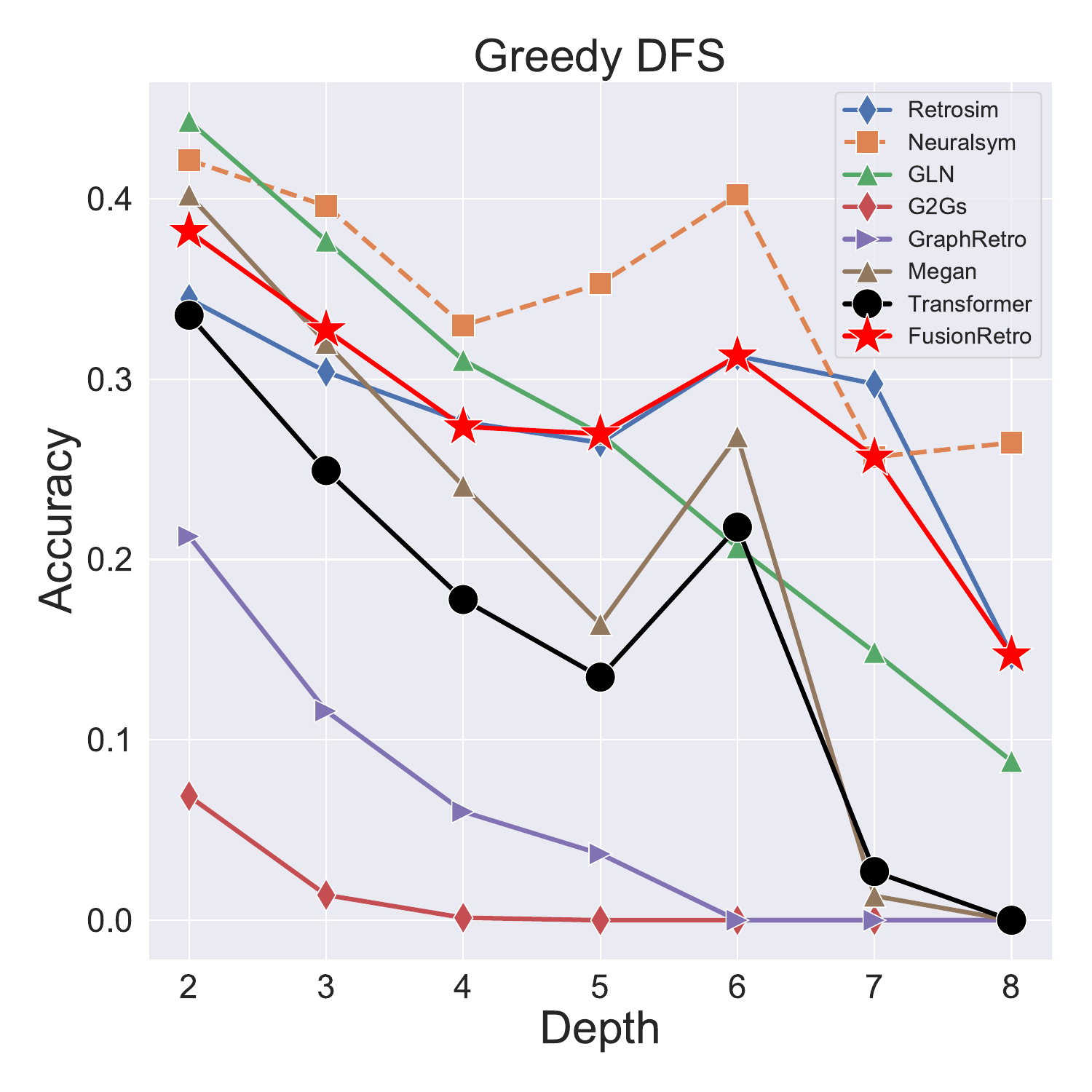}
    \end{center}
    \vskip -0.2in
    \caption{The test accuracy of retrosynthesis models combined with Greedy DFS at different depths of the ground truth synthetic routes. \textit{Red stars} (\textcolor{red}{\ding{72}}) denotes our method (FusionRetro) and \textit{Black circles} ($\textcolor{black}{\bullet}$) represents Transformer.}
    \label{fig:depth}
    \vskip -0.1in
 \end{figure}

\paragraph{Comparison with Template-free Baselines.}The primary results are presented in Table~\ref{tab:main results}. It's clear that our proposed model, \name, outperforms other template-free baseline methods. Further insights can be drawn from Figure~\ref{fig:depth}, which shows that as the depth of the ground truth synthetic routes increases, the performance gap between the Transformer and \name generally widens. This demonstrates the value of incorporating context information for representation fusion. In essence, these results indicate that our proposed model consistently performs better than Transformer, particularly in predicting long synthetic routes.

\paragraph{Analysis of the Benchmark.} The performance of baseline models on our benchmark does not align well with single-step retrosynthesis predictions on the USPTO-50K dataset. Current two-stage semi-template-based models~\citep{shi2020graph,somnath2021learning} either outperform or match template-based and template-free models on USPTO-50K single-step retrosynthesis prediction, yet perform poorly on our benchmark. One main factor is that approximately 95\% of reactions in the USPTO-50K dataset have only one reaction center due to heavy filtering, whereas in our constructed dataset, around 30\% of reactions have multiple reaction centers. Upon examining the open-source code of G2Gs, we found that it can only handle cases with one reaction center, which explains its weak performance on our benchmark. The performance of template-free models is not impacted by the number of reaction centers. Additionally, we present the results of single-step retrosynthesis predictions on our constructed test dataset in Table~\ref{tab:retrosynthesis}. These results align with those of retrosynthetic planning in Table~\ref{tab:main results}, leading us to conclude that single-step accuracy plays a crucial role in multi-step planning as well.

\paragraph{Analysis for the Depth of Routes.} As illustrated in Figure~\ref{fig:depth}, the accuracy of prediction tends to decrease as the depth of synthetic routes increases. However, our model exhibits a slower rate of performance degradation compared to other baseline models. This indicates the strength of our approach, which uses contextual information for representation fusion, particularly when predicting long synthetic routes.

\begin{table}[t]
\centering
\caption{Summary of retrosynthesis prediction results in terms of exact match accuracy (\%).}
\label{tab:retrosynthesis}
\vskip 0.1in
\setlength{\tabcolsep}{4.0mm}
\begin{tabular}{ccccc}
        \toprule
     \multirow{2}{*}{Methods}   & \multicolumn{4}{c}{Top-$k$ accuracy \%} \\
        \cmidrule{2-5}
         & 1 & 3 & 5 & 10 \\
        \midrule
         G2Gs & 16.5 & 27.8 & 33.1 & 40.4\\
         GraphRetro & 48.3 & 58.4 & 60.5 & 62.4  \\
         Transformer & 55.8 & 70.3 & 74.8 & 78.9 \\ 
         Retrosim & 56.5 & 65.8 & 69.0 & 73.1 \\
         Megan & 59.5 & 73.9 & 77.9 & 81.7 \\
         Neuralsym & \textbf{63.0} & 73.3 & 76.0 & 78.6 \\
         GLN & 62.9 & \textbf{74.1} & \textbf{78.4} & \textbf{82.7} \\
		 \bottomrule
    \end{tabular}
    \vskip -0.1in
\end{table}

\begin{figure}[t]
    \begin{center}
        \includegraphics[width=0.48\textwidth]{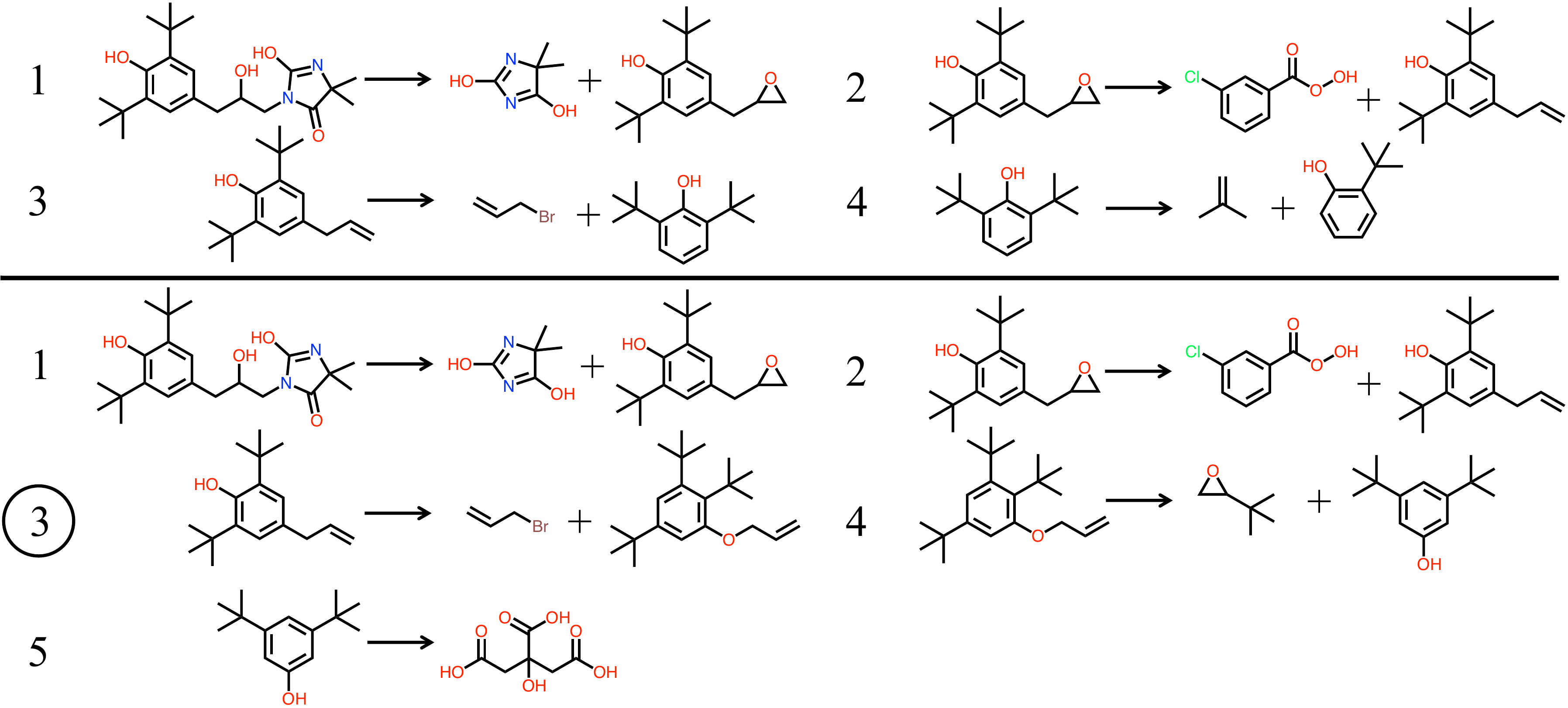}
    \end{center}
    \caption{We split the predicted synthetic route into individual reactions. The correct synthetic route predicted by \name is depicted at the top, while the route predicted by the Transformer model is displayed at the bottom.}
   \label{fig:case_study}
    \vskip -0.1in
\end{figure}

\subsection{Case Study}
Figure~\ref{fig:case_study} provides a visual comparison of predictions made by \name and Transformer. The upper section of the figure displays accurate predictions made by \name, while the lower section shows incorrect predictions made by Transformer. It is evident from the figure that Transformer inaccurately predicts the third-step retrosynthesis reaction. Although a search can still identify starting materials, these materials may not be capable of synthesizing the target molecule. Therefore, the performance of retrosynthesis prediction is crucial for effective retrosynthetic planning.

\section{Conclusion and Future Work}
In this paper, we propose \name, a novel framework for retrosynthetic planning that exploits crucial context information on the synthetic route by principled representation fusion. \name is the first method in this field that takes context information into account, greatly boosting the performance for realistic multi-step planning. We further introduce new benchmarks for better evaluation of retrosynthesis models, especially for practical multi-step planning settings. Extensive experiments demonstrate \name can consistently achieve significantly superior performance across several measurements. We hope our approach can shed light on the research of data-driven retrosynthetic planning, and inspire more studies toward the practical multi-step scenario. Besides, our approach can be viewed as in-context learning and can inspire more works to further explore in-context learning techniques in large language models for multi-step planning. In this way, we can enrich the decision-making process with valuable context-driven inputs.

\section*{Acknowledgements}
We thank all the anonymous reviewers and area chairs for their helpful comments and suggestions. Songtao Liu thanks Binghong Chen, Hanjun Dai, Samuel Genheden, Connor W. Coley, Tianfan Fu, Chenghao Yang, Peng Han, Guoren Xi, Shen Yuan, Yang Yu, Ziqiao Meng, Chan Lu, Changrui Fan, Jiatong Li, and Yunhua Zhou for their helpful discussions and comments. Minkai Xu thanks the generous support of Sequoia Capital Stanford Graduate Fellowship.


\bibliography{icml2023}
\bibliographystyle{icml2023}

\newpage
\appendix
\onecolumn
\section{Datasets Details}
\label{appendix:dataset}
\begin{table}[htbp]
\centering
\caption{The number of target molecules in training/validation/test datasets in term of the shortest depths to synthesize the target molecules.}
\vspace*{\baselineskip}
\label{tab:stat}
\setlength{\tabcolsep}{1mm}
\begin{tabular}{lcccccccccccc}
\toprule
  \diagbox{\textbf{Dataset}}{\textbf{\#Molecules}}{\textbf{Depth}}     & 2  &3 & 4 & 5  & 6   & 7 & 8 & 9 & 10 & 11 &12 & 13   \\
\midrule
Training    & 22,903   & 12,004    & 5,849    &3,268 & 1,432    & 594     & 276     & 107 & 25 & 0 & 0 & 0   \\
Validation    & 2,862   & 1,500    & 731    &408 & 179    & 74     & 34     & 13 & 2 & 0 & 0 & 0   \\
Test    & 2,862   & 1,500    & 731    &408 & 179    & 74     & 34     & 13 & 2 & 32 & 2 & 1   \\
\bottomrule
\end{tabular}
\end{table}

\section{Reproducibility}
\subsection{Implementation Details}
\label{appendix:implementation}
We use Pytorch~\citep{paszke2019pytorch} to implement \name. The codes of baselines are implemented referring to the implementation of Retrosim\footnote{https://github.com/connorcoley/retrosim}, Neuralsym\footnote{https://github.com/linminhtoo/neuralsym}, GLN\footnote{https://github.com/Hanjun-Dai/GLN}, G2Gs\footnote{https://torchdrug.ai/docs/tutorials/retrosynthesis}, GraphRetro\footnote{https://github.com/vsomnath/graphretro}, Transformer\footnote{https://github.com/bigchem/synthesis}, and Megan\footnote{https://github.com/molecule-one/megan}. All the experiments of baselines are conducted on a single NVIDIA Tesla V100 with 32GB memory size. The software that we use for experiments are Python 3.6.8, pytorch 1.9.0, pytorch-scatter 2.0.9, pytorch-sparse 0.6.12, numpy 1.19.2, torchvision 0.10.0, CUDA 10.2.89, CUDNN 7.6.5, einops 0.4.1, and torchdrug 0.1.3.

\subsection{Hyperparameter Details}
\begin{table}[htbp]
\setlength{\tabcolsep}{0.3mm}
\caption{The hyper-parameters for \name.}
\label{tab:heterophily_hyper}
\vskip 0.15in
    \centering
    \begin{tabular}{l|c}
    \toprule
     max length & 200\\
     embedding size & 64\\
     encoder layers & 3\\
     decoder layers & 3\\
     fusion layers & 3\\
     attention heads & 10\\
     FFN hidden & 512\\ 
     dropout & 0.1\\
     epochs & 4000\\
     batch size & 64\\
     warmup & 16000\\
     lr factor & 20\\
    \bottomrule
    \end{tabular}
\end{table}

\section{More results}

 \begin{figure}[t]
    \begin{center}
        \includegraphics[width=0.4\textwidth]{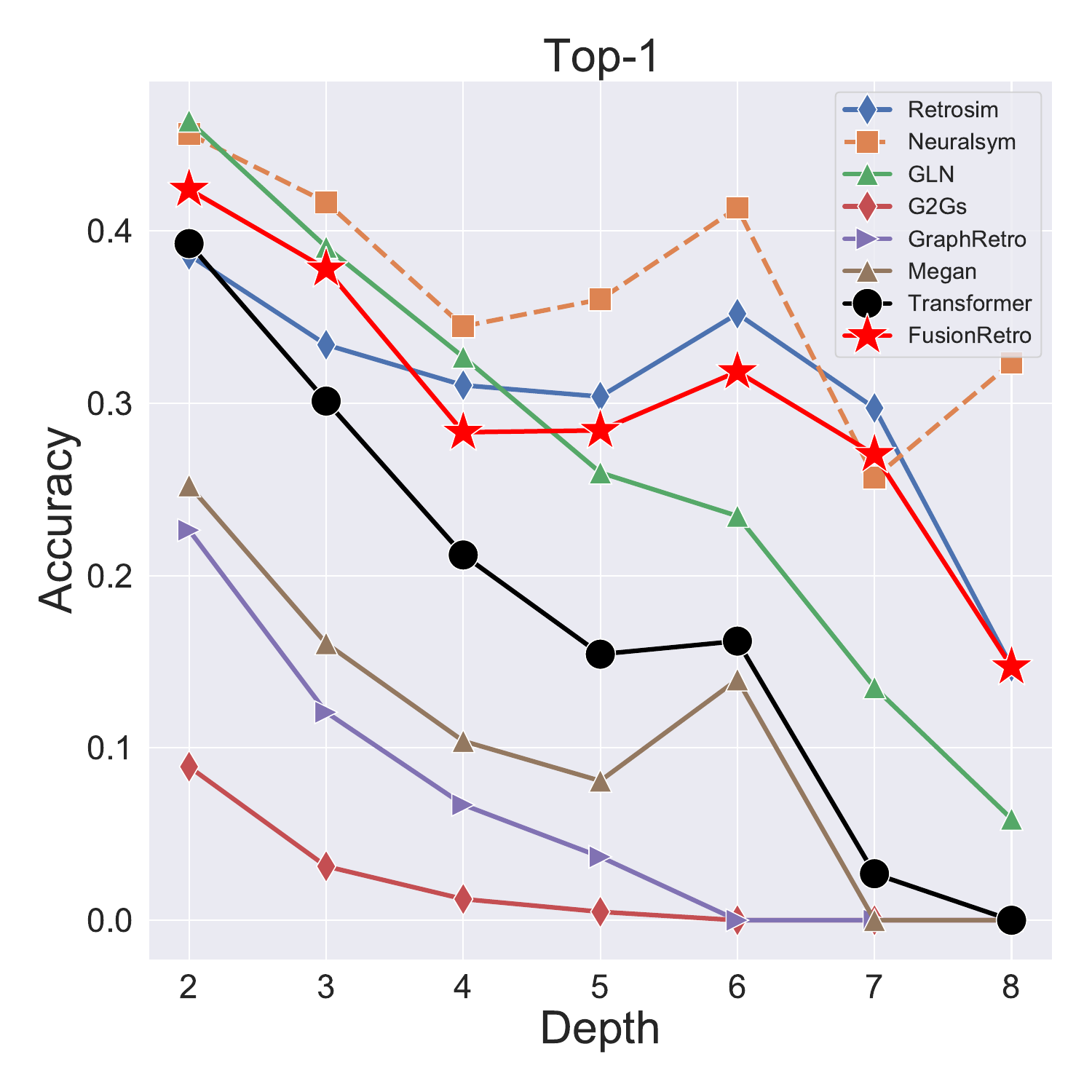}
        \includegraphics[width=0.4\textwidth]{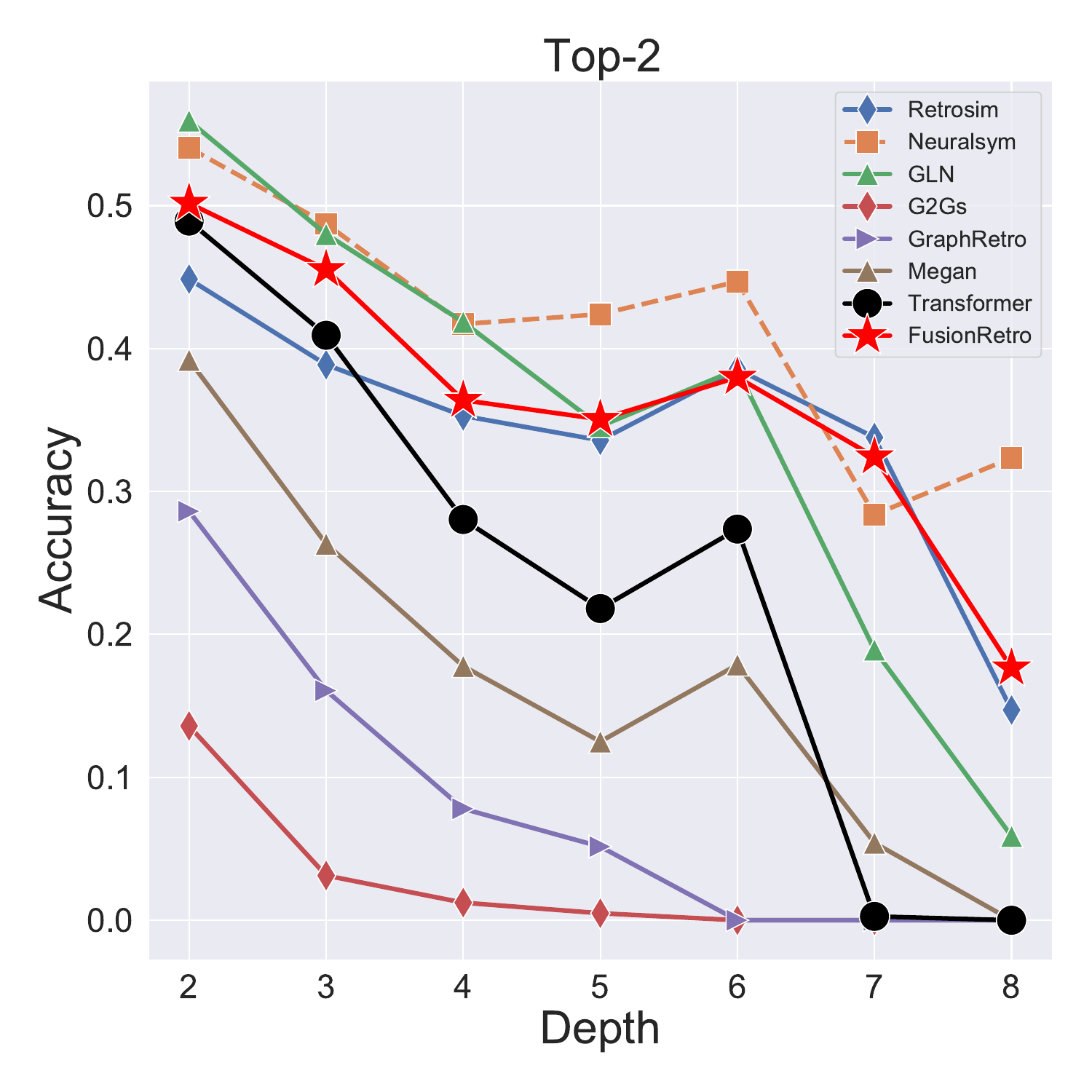}
        \includegraphics[width=0.4\textwidth]{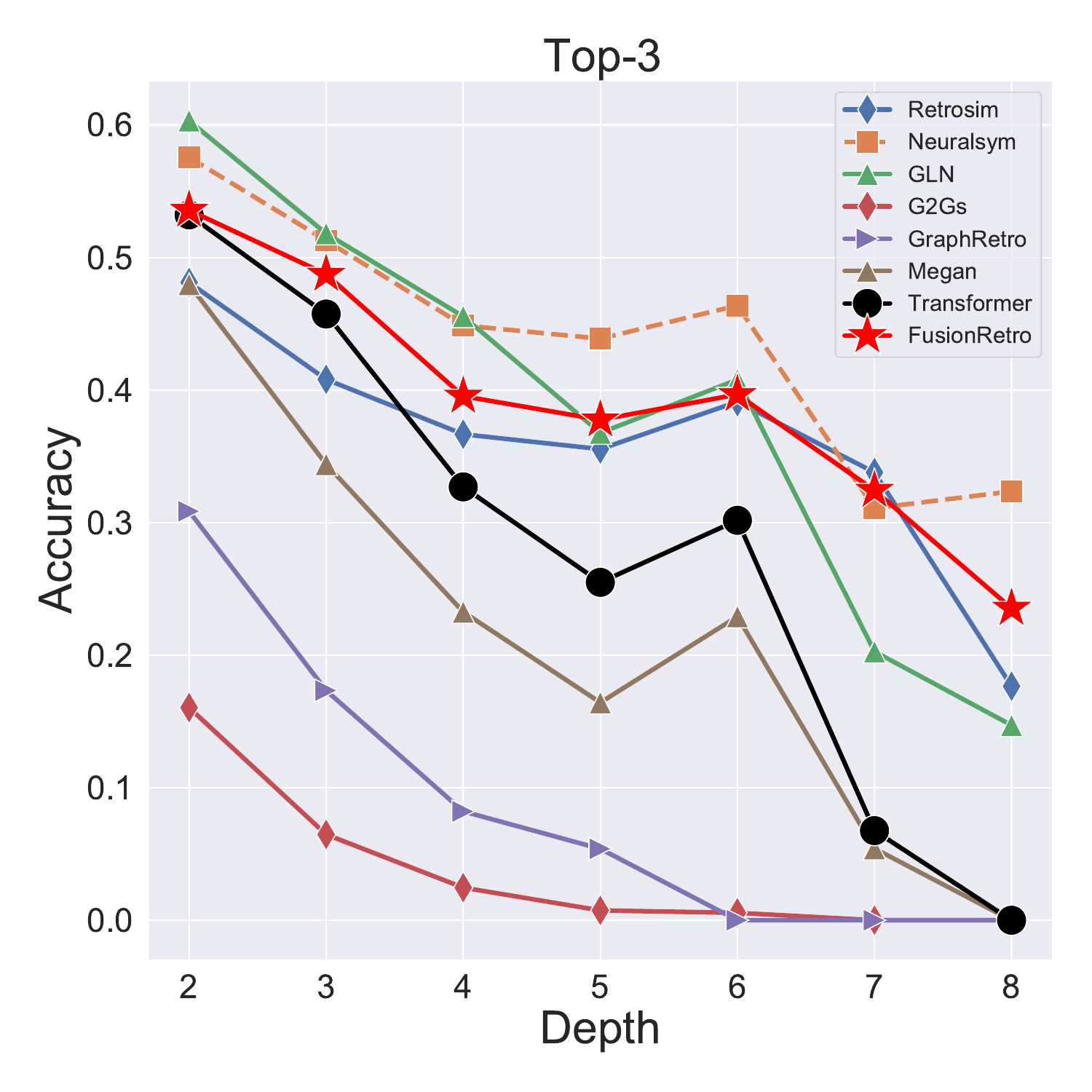}
    \end{center}
    \vspace*{-\baselineskip}
    \caption{The top-1, top-2, and top-3 test accuracy in terms of depth.}
 \end{figure}
 
 \begin{figure}[t]
    \begin{center}
        \includegraphics[width=0.4\textwidth]{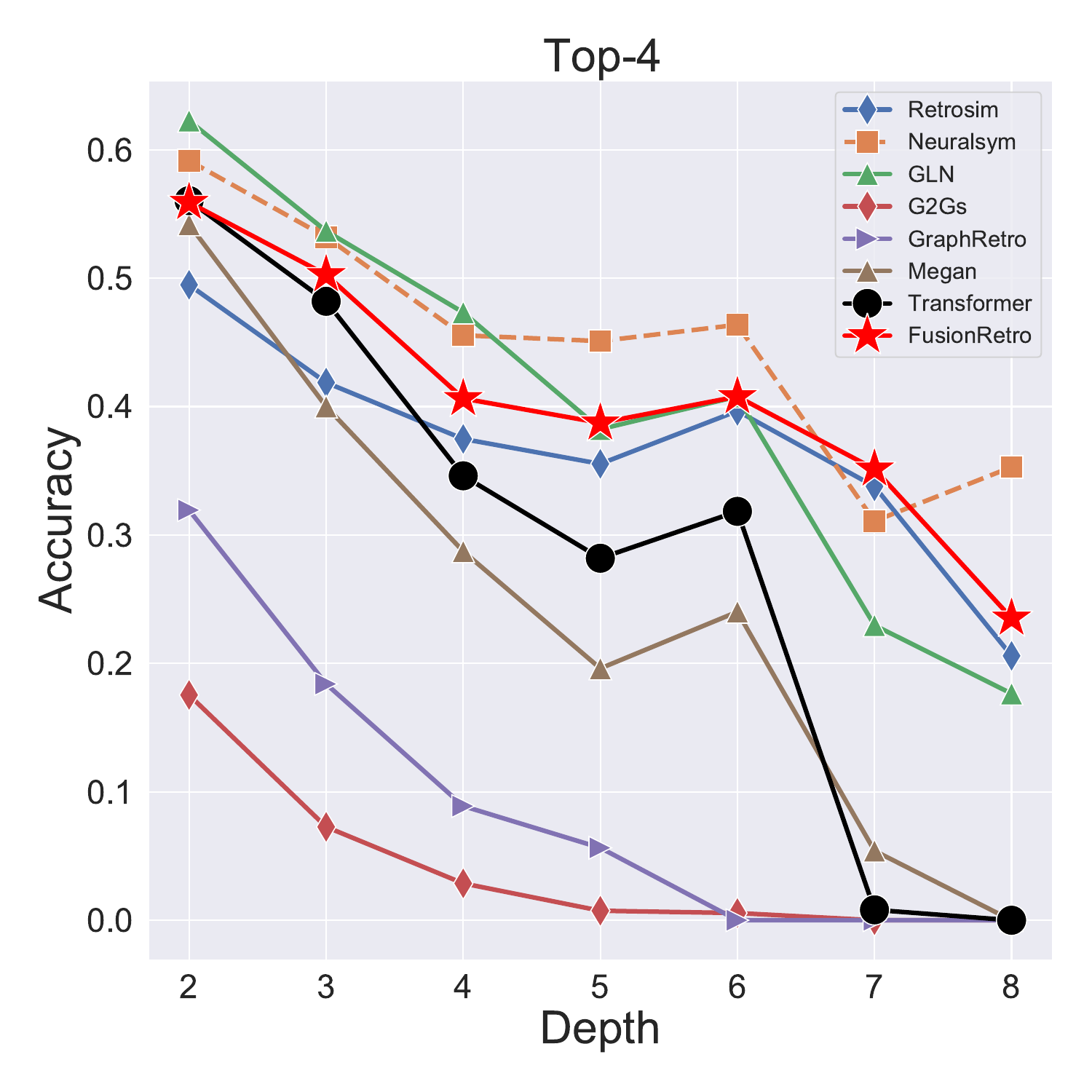}
        \includegraphics[width=0.4\textwidth]{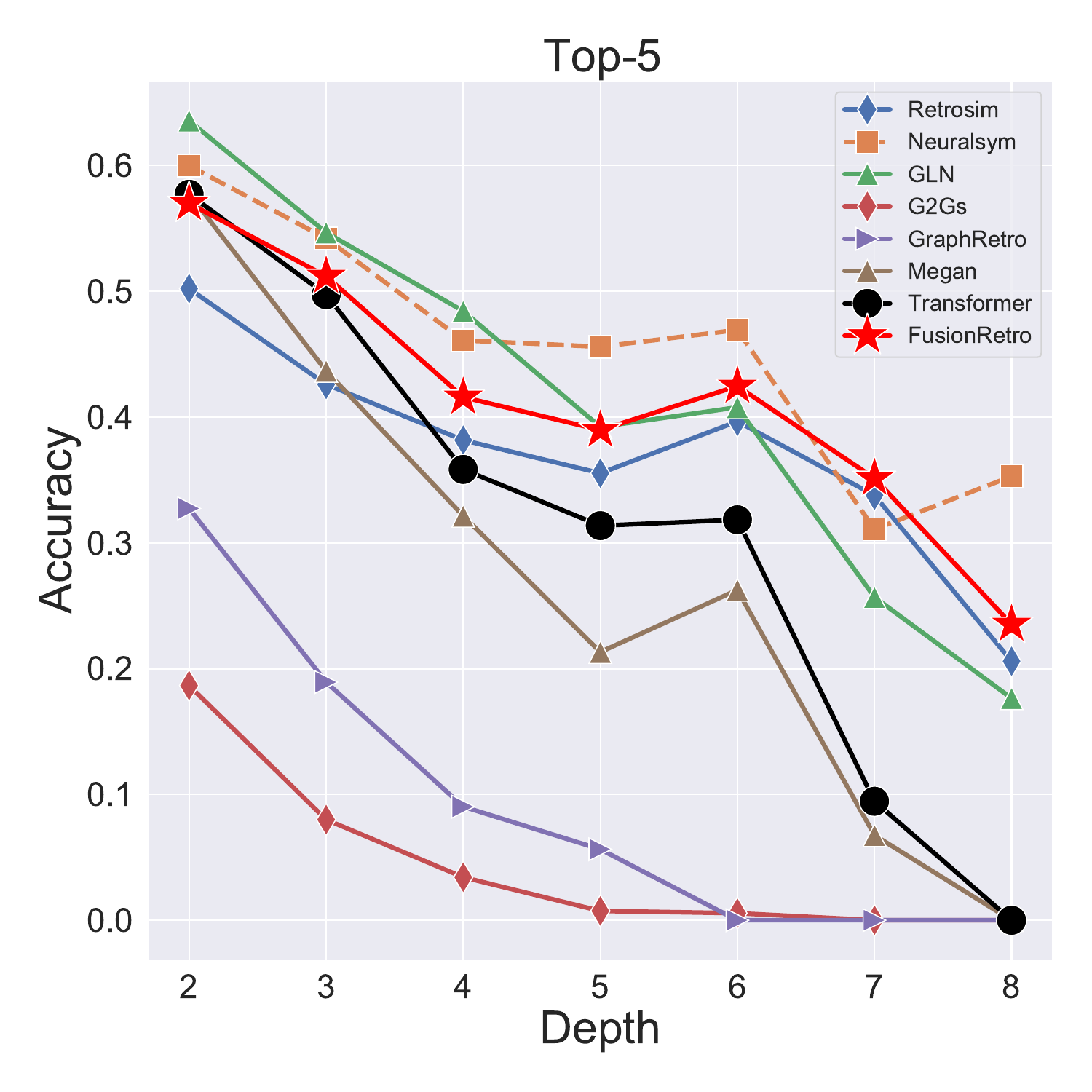}
    \end{center}
    \vspace*{-\baselineskip}
    \caption{The top-4 and top-5 test accuracy in terms of depth.}
 \end{figure}

\end{document}